\documentclass[sigconf]{acmart}
\AtBeginDocument{%
  \providecommand\BibTeX{{%
    Bib\TeX}}}

\usepackage{pifont}
\newcommand{\cmark}{\ding{51}} 
\newcommand{\xmark}{\ding{55}} 

\usepackage{xcolor}

\usepackage[linesnumbered,ruled]{algorithm2e}
\usepackage{algpseudocode}
\usepackage{algorithmicx}  

\usepackage{amsmath}

\usepackage{amssymb}
\usepackage{multirow} 
\usepackage{tabularx}
\newcolumntype{C}{>{\centering\arraybackslash}X}
\usepackage{marvosym}
\usepackage{colortbl}
\usepackage{makecell}
\usepackage{booktabs}

\newcommand{\xf}[1]{{\color{blue}#1}}

\setcopyright{none}
\renewcommand\footnotetextcopyrightpermission[1]{}

\makeatletter
\def\@affiliationfont{\small\normalfont}
\let\@ACM@checkaffil\relax
\makeatother

\begin{document}

\title{Detector-Empowered Video Large Language Model for Efficient Spatio-Temporal Grounding}

\author{%
Shida Gao$^{1,*}$ \quad
Feng Xue$^{2,*}$ \quad
Xiangfeng Wang$^{1,*}$ \quad
Anlong Ming$^{1,\dagger}$ \quad
Zhaowen Lin$^{1}$ \\
Haiyang Zhang$^{1}$ \quad
Teng Long$^{2}$ \quad
Nicu Sebe$^{2}$ \quad
Yihua Shao$^{3}$ \quad
Haozhe Wang$^{4}$ \quad
Wei Wang$^{5}$}
\affiliation{%
    \institution{%
    \parbox{\hsize}{\centering
    $^{1}$Beijing University of Posts and Telecommunications \quad
    $^{2}$University of Trento \quad
    $^{3}$Institute of Automation, Chinese Academy of Sciences \\
    $^{4}$Hong Kong University of Science and Technology \quad
    $^{5}$ZTE Corporation \\
    \textbf{Code and dataset:} \href{https://github.com/gaostar123/DeViL}{\textcolor{blue}{https://github.com/gaostar123/DeViL}}}
    }
}

\renewcommand{\shortauthors}{Gao et al.}

\begin{abstract}
Multimodal large language models (MLLMs) are rapidly expanding from general video understanding to finer-grained understanding such as spatio-temporal video grounding (STVG) and reasoning.
In these tasks, an MLLM must localize the user-queried target in time and space and take the results as evidence for reasoning.
Existing MLLM methods mainly follow two paradigms.
\ding{172} Direct Localization, which outputs STVG results with extra alignment modules or specialized decoders.
\ding{173} Candidate-based Selection,
which first constructs tube-level candidates and then select the relevant one by an MLLM.
However, both suffer from a \textit{serious efficiency bottleneck}:
the former incurs linear-growing decoding cost as queried temporal span increases,
while the latter relies on costly candidate construction.
To break this bottleneck, we propose DEViL,
a detector-empowered Video-LLM with a simple key idea:
\textit{offloading dense spatial grounding from the MLLM to a fully-parallelizable, well-trained detector}.
Specifically, DEViL distills the query into a detector-compatible reference-semantic token,
which replaces the detector’s text embedding to enable spatial grounding in a single pass.
Then, we design temporal consistency regularization to match objects across frames and enforce their coherence over time.
In this way,
DEViL avoids long coordinate decoding and heavy candidate pipelines. Extensive experiments show that DEViL achieves strong performance ($43.1\%$ m\_vIoU on HC-STVG) with superior efficiency ($14.33$ FPS),
while preserving the general reasoning capacity of the MLLM backbone.
\end{abstract}


\begin{teaserfigure}
  \includegraphics[width=\textwidth]{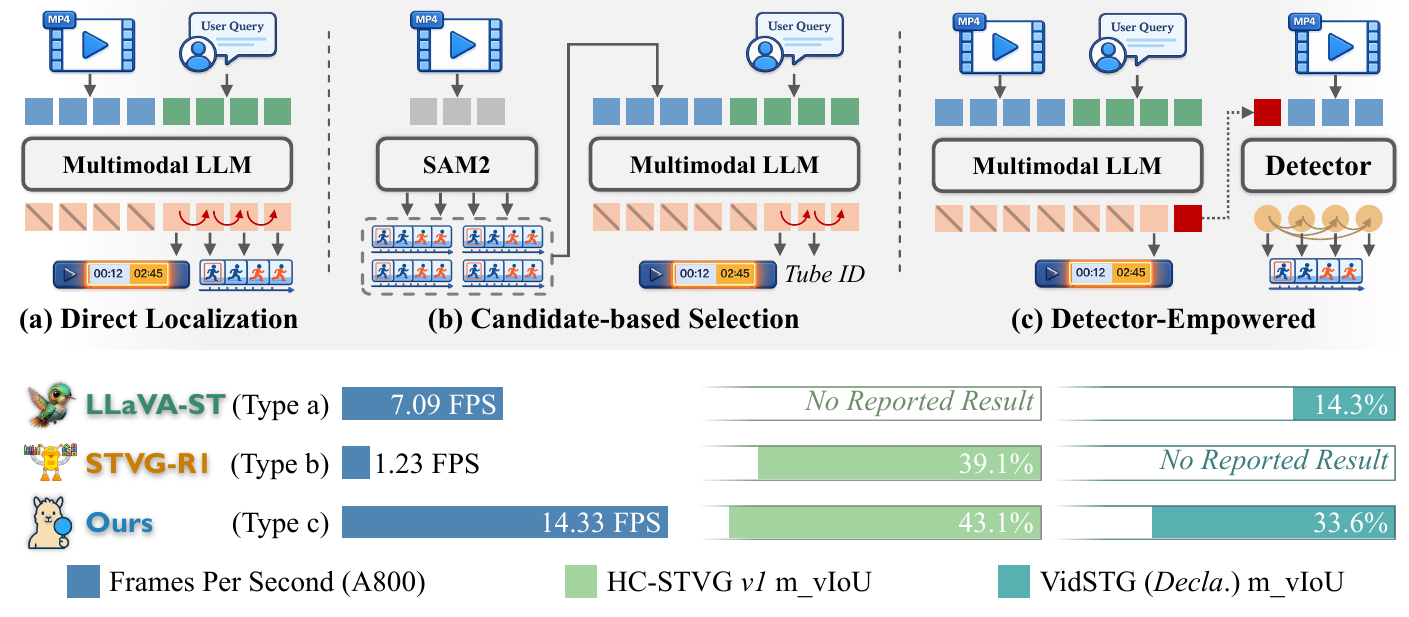}
\caption{\textbf{Comparison of spatio-temporal video grounding paradigms.} Comparison among (a) \textbf{LLaVA-ST}~\cite{llavast}, which directly predicts spatio-temporal grounding outputs from the MLLM, (b) \textbf{STVG-R1}~\cite{zhang2026stvg}, which first constructs object tubes and then lets the MLLM select the relevant one, and (c) our \textbf{DEViL}, which directly couples an MLLM with an open-vocabulary detector. DEViL achieves superior spatio-temporal grounding performance while maintaining high inference efficiency.}
  \label{fig:teaser}
\end{teaserfigure}

\maketitle
\begingroup
\renewcommand{\thefootnote}{}
\footnotetext{$^{*}$Equal contribution. $^{\dagger}$Corresponding author.}
\addtocounter{footnote}{-1}
\endgroup

\label{sec:intro}

\section{Introduction}
Driven by the recent progress of multimodal large language models (MLLMs),
video understanding has rapidly progressed from \textbf{coarse-grained} reasoning tasks,
such as video question answering (VQA) \cite{vqa,vqa2}, captioning (VC) \cite{xu2016msrvtt,chen2011msvd}, and summarization (VS) \cite{song2015tvsum,gygli2014summe},
toward \textbf{fine-grained} spatio-temporal reasoning,
including video kinematics \cite{yi2020clevrer} and grounded VQA (GVQA) \cite{xiao2024can,lei2020tvqa_plus,gan2023temporal,liu2022reducing}.
With this shift, spatio-temporal video grounding (STVG) serves as a natural bridge between coarse video understanding and fine-grained spatio-temporal reasoning.
STVG requires locating both the temporal segment and the spatial trajectory (frame-wise bounding boxes) of the object mentioned in the user query.

Recent studies have endowed MLLMs with spatio-temporal video grounding \cite{RealVG,wang2023efficient,wang2023deconfounded}, yielding remarkable results on STVG and related downstream tasks. According to their inference structure, existing MLLM methods generally fall into two paradigms, as compared in Fig. \ref{fig:teaser}.
The first, \textit{e.g.}, LLaVA-ST \cite{llavast}, SpaceVLLM \cite{spacevllm}, Open-o3-Video \cite{openo3video}
directly predicts STVG outputs by MLLMs, often with additional alignment modules,
temporal queries, or specialized decoders to improve fine-grained coordinate prediction.
The second type, \textit{e.g.}, STVG-R1 \cite{zhang2026stvg},
first constructs tube-level candidates and then letting the MLLM select the most relevant ones according to the query.
However, both of them suffer from substantial efficiency bottlenecks. Direct localization methods often require dense autoregressive decoding or explicit intermediate localization steps,
causing inference cost to grow rapidly with video length and grounding granularity.
Candidate-based methods, while avoiding direct coordinate generation,
still rely on expensive pre-processing before reasoning can even begin.
Consequently, existing MLLMs with STVG ability remain difficult to scale efficiently to realistic settings,
where both temporal span and visual complexity are large.


In this work, we break the efficiency bottleneck with a detector-empowered Video-LLM method, short for DEViL.
Instead of forcing the MLLM to directly decode dense spatial locations or rely on heavy pre-built candidate pipelines,
DEViL delegates dense spatial localization to a fully parallelizable detector, while keeping sparse temporal grounding within the MLLM.
This decomposition brings three key advantages:
it improves efficiency, requires only minimal modification to the MLLM,
and directly inherits the mature spatial localization capability of a well-trained detector.
Specifically, we first use the MLLM to transform the user query into a detector-compatible reference-semantic token (RST), which replaces the detector’s text embedding and enables frame-wise spatial ground in parallel.
To associate independent detections across frames,
we then design a temporal consistency regularization that makes only minimally invasive modifications to the detector and turns it into a tunable tracker.
This regularization can be applied in a single pass, enabling fast inference.
In this way, DEViL achieves a better balance between grounding accuracy and efficiency.
Extensive experiments show that DEViL delivers stronger spatio-temporal grounding performance while running significantly more efficiently than previous MLLMs.
In summary, the major contributions of our work are:

\begin{itemize}
\item We rethink MLLM-based STVG from an efficiency perspective and propose a new decomposition that assigns sparse temporal grounding to the MLLM and dense spatial localization to a fully parallelizable detector.
\item We show that efficient spatio-temporal grounding does not require heavily specializing the MLLM for dense localization, but can instead be achieved by tightly integrating query understanding in the MLLM with the mature perception capability of a well-trained detector.
\item This design brings both higher efficiency (\textbf{\textit{near 15 FPS}}) and stronger grounding performance (\textbf{\textit{43.1\% m\_vIoU on HC-STVG v1 and 33.6\% m\_vIoU on VidSTG Decla.}}),
outperforming prior STVG-oriented MLLMs while keeping generalization to conventional video understanding.
\end{itemize}

\section{Related Works}
\label{sec:related_works}

\subsection{Multimodal LLMs for Video Understanding}
Current MLLMs mainly target coarse-grained video understanding tasks, such as Video Question Answering (VQA) \cite{grauman2022ego4d,zhang2023video,song2024moviechat,2023videochat, maaz2024video}, and are gradually being extended toward finer-grained settings, including Temporal Video Grounding (TVG) \cite{anne2017localizing,guo2025vtg,huang2024vtimellm,huang2024lita,guo2024trace,wang2023mixup,jiang2024counterfactually,woo2024let}, Referring Video Object Segmentation (RVOS) \cite{yan2024visa,videolisa,lin2025glus,Sa2va,yan2024tracking}, and Grounded VQA (GVQA) \cite{grunde2021agqa,qian2024momentor,wang2024grounded}. A few recent works further explore Spatio-Temporal Video Grounding (STVG) within the MLLM framework \cite{llavast,openo3video,spacevllm,zhang2026stvg}. However, methods based on textualized coordinate decoding often require dense autoregressive prediction, causing inference cost to grow rapidly with video length and grounding granularity, while decoupled ``segment-then-select'' paradigms (\textit{e.g.}, STVG-R1 \cite{zhang2026stvg}) rely on rigid pre-extracted candidates, which bottleneck spatial precision and incur substantial inference latency. To address these limitations, DEViL replaces the detector's original text embedding with a learned Reference-Semantic Token (RST), enabling efficient one-pass spatial grounding through parallel detector perception. Combined with temporal reasoning in the MLLM, this design produces temporally consistent tubes while preserving broad video understanding capability.

\begin{figure*}[t]
\centering
\includegraphics[width=1\linewidth]{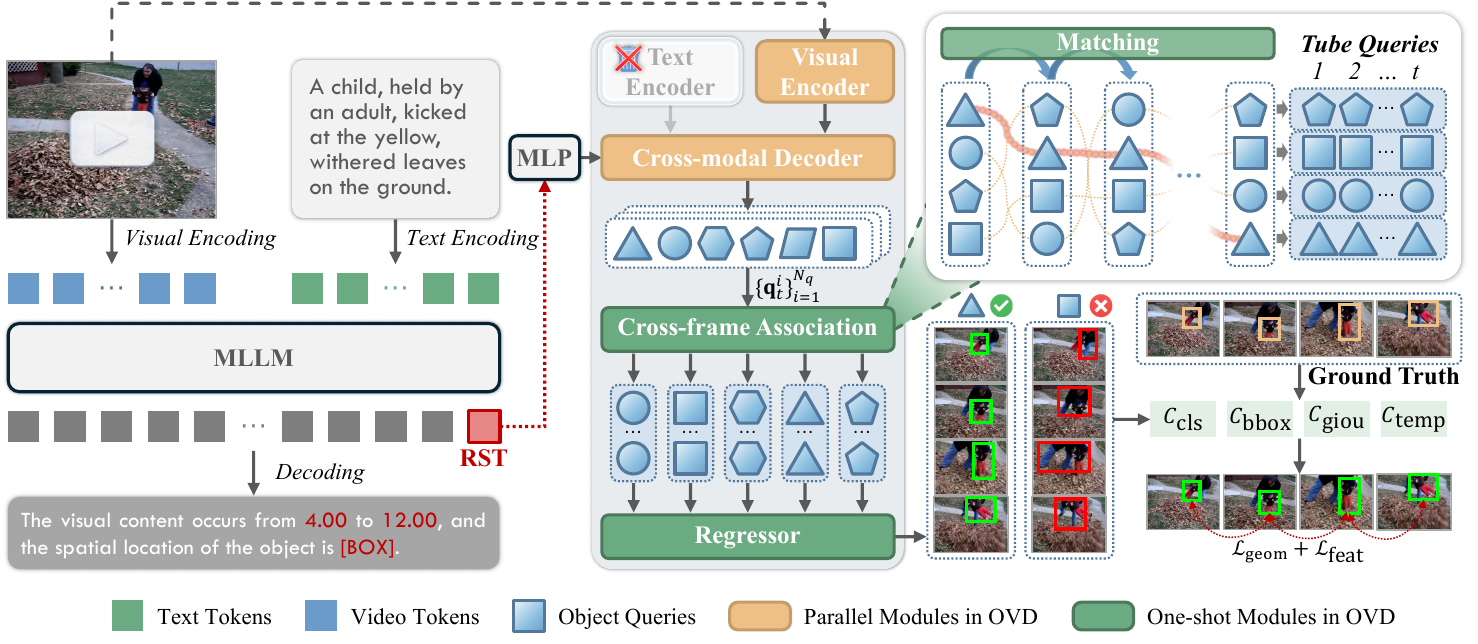}
\caption{Overall architecture of DEViL.
Given a video and query,
the MLLM encodes them and emits a special \texttt{[BOX]} token whose hidden state serves as the Reference-Semantic Token (RST).
RST replaces the text embedding of the open-vocabulary detector (OVD) to drive object queries.
A memory-based tube association maintains query identity across frames,
while tube-mined temporal regularization (TTReg) regularizes ground-truth–aligned tubes to learn temporally consistent boxes.
Note that the classification head of OVD is omitted for the purpose of simplifying expression and visualization.}
\label{fig:pipeline}
\end{figure*}

\subsection{Spatial-Temporal Grounding and Reasoning}
Recent MLLM-based approaches to video evidence localization can be roughly grouped into three types.
First, \emph{video spatial grounding} models predict frame-wise masks or boxes under language guidance while treating time implicitly through tracking \cite{yan2024visa,videolisa,lin2025glus,Sa2va}, and therefore usually do not output explicit start--end timestamps or complete what--when--where grounding results.
Although structurally related, DEViL differs from these methods by using a learned RST to align the MLLM with an open-vocabulary detector for tube-level grounding, rather than using prompt tokens to trigger a segmentor.
Second, \emph{video temporal grounding} methods align queries with temporal segments \cite{qian2024momentor,huang2024vtimellm,ren2024timechat,guo2025vtg,wang2024hawkeye,wang2024grounded}, but their spatial localization is typically coarse and not modeled as frame-wise tubes.
Third, \emph{spatio-temporal video grounding} methods within the MLLM framework aim to predict both when and where \cite{llavast,openo3video,spacevllm,zhang2026stvg,tian2025ddavs}. However, existing methods still face substantial efficiency bottlenecks: textualized coordinate decoding scales poorly due to dense autoregressive prediction, while decoupled ``segment-then-select'' methods rely on rigid pre-extracted candidates, increasing inference latency and limiting spatial precision. In contrast, DEViL couples an MLLM with an open-vocabulary detector through the learned RST and further applies tube-mined temporal regularization (TTReg) to improve cross-frame consistency. This design allows DEViL to jointly ground both temporal evidence and spatial trajectories in a more efficient way.

\section{Method}
\label{sec:method}


\subsection{Overview}
\label{sec:overview}

The overall architecture of DEViL is illustrated in Fig.~\ref{fig:pipeline}.
DEViL consists of two components:
a multimodal large language model (MLLM) and an open-vocabulary detector (OVD).
Given an input video and a text query,
the MLLM predicts when the queried event occurs,
while the OVD localize the target frame-wisely.
To connect language reasoning with detector-based perception,
DEViL uses a specialized reference-semantic token (RST) generated by the MLLM (see Sec. \ref{sec:spatial_grounding}).
Conditioned on this token, the OVD performs query-aware frame-level localization,
while a subsequent temporal consistency regularization module associates frame-wise detections into smooth and accurate trajectories (see Sec. \ref{sec:tce}).
In this way, DEViL forms an efficient and scalable pipeline for spatio-temporal grounding and reasoning.



\subsection{Reference Semantic-conditioned Grounding}
\label{sec:spatial_grounding}
To bridge the MLLM and the OVD, DEViL introduces a reference-semantic token (RST),
which transfers query-conditioned referential semantics from the MLLM to the detector for spatial localization.
This design is efficient and minimally invasive:
the MLLM remains focused on language reasoning and temporal grounding, while the detector handles dense spatial perception in parallel.

\subsubsection{Reference-Semantic Token Generation.}
Given a video $\{v_t\}_{t=1}^{T}$ with $T$ frames and a textual query $Q$,
the MLLM processes them and produces a sequence of hidden states:
\begin{equation}
    \mathbf{H}_{\mathrm{llm}} \in \mathbb{R}^{L \times D_{\mathrm{llm}}},
\end{equation}
where $L$ is the number of output tokens and $D_{\mathrm{llm}}$ is the hidden dimension.
When spatial localization is required, the MLLM is trained to emit a special token, denoted as $\texttt{[BOX]}$. The hidden state of this token is taken as the Reference-Semantic Token,
\begin{equation}
    \mathbf{z}_{\mathrm{rst}} \in \mathbb{R}^{D_{\mathrm{llm}}},
\end{equation}
which encodes the referential semantics needed for query-conditioned spatial localization.

\subsubsection{RST-conditioned Detector Grounding.}
To make the RST compatible with the detector, we project $\mathbf{z}_{\mathrm{rst}}$ into the detector text-embedding space using a learnable linear layer:
\begin{equation}
    \mathbf{e}_{\mathrm{text}} = \mathbf{W}\mathbf{z}_{\mathrm{rst}} + \mathbf{b},
    \quad
    \mathbf{e}_{\mathrm{text}} \in \mathbb{R}^{D_{\mathrm{det}}},
    \label{eq:rst_projection}
\end{equation}
where $\mathbf{W}\in\mathbb{R}^{D_{\mathrm{det}}\times D_{\mathrm{llm}}}$ and $\mathbf{b}\in\mathbb{R}^{D_{\mathrm{det}}}$ are learnable parameters. We then use $\mathbf{e}_{\mathrm{text}}$ to replace the detector's original text embedding, feeding it into the OVD as the language-side query for cross-modal interaction.
Conditioned on this shared query representation, the OVD performs query-aware spatial localization for all frames in parallel. Specifically, for each frame $t$, the detector decoder produces a set of query features and their corresponding box predictions:
\begin{equation}
    \Big\{\{(\mathbf{q}_t^i, b_t^i)\}_{t=1}^{T}\Big\}_{i=1}^{N}
    =
    \mathtt{OVD}\!\left(\{v_t\}_{t=1}^{T}, \mathbf{e}_{\mathrm{text}}\right).
\end{equation}
where $\mathbf{q}_t^i$ denotes the $i$-th decoder query and $b_t^i$ is its predicted bounding box. In this way, DEViL avoids expensive autoregressive spatial decoding in the MLLM and instead delegates dense localization to a detector that is naturally suited for parallel perception. At the same time, the MLLM only needs to produce a compact semantic interface, which keeps the overall design efficient and lightweight.
These frame-wise query outputs further serve as the basis for the temporal consistency regularization introduced in Sec.~\ref{sec:tce}.

\begin{figure}[t]
\centering
\includegraphics[width=1\linewidth]{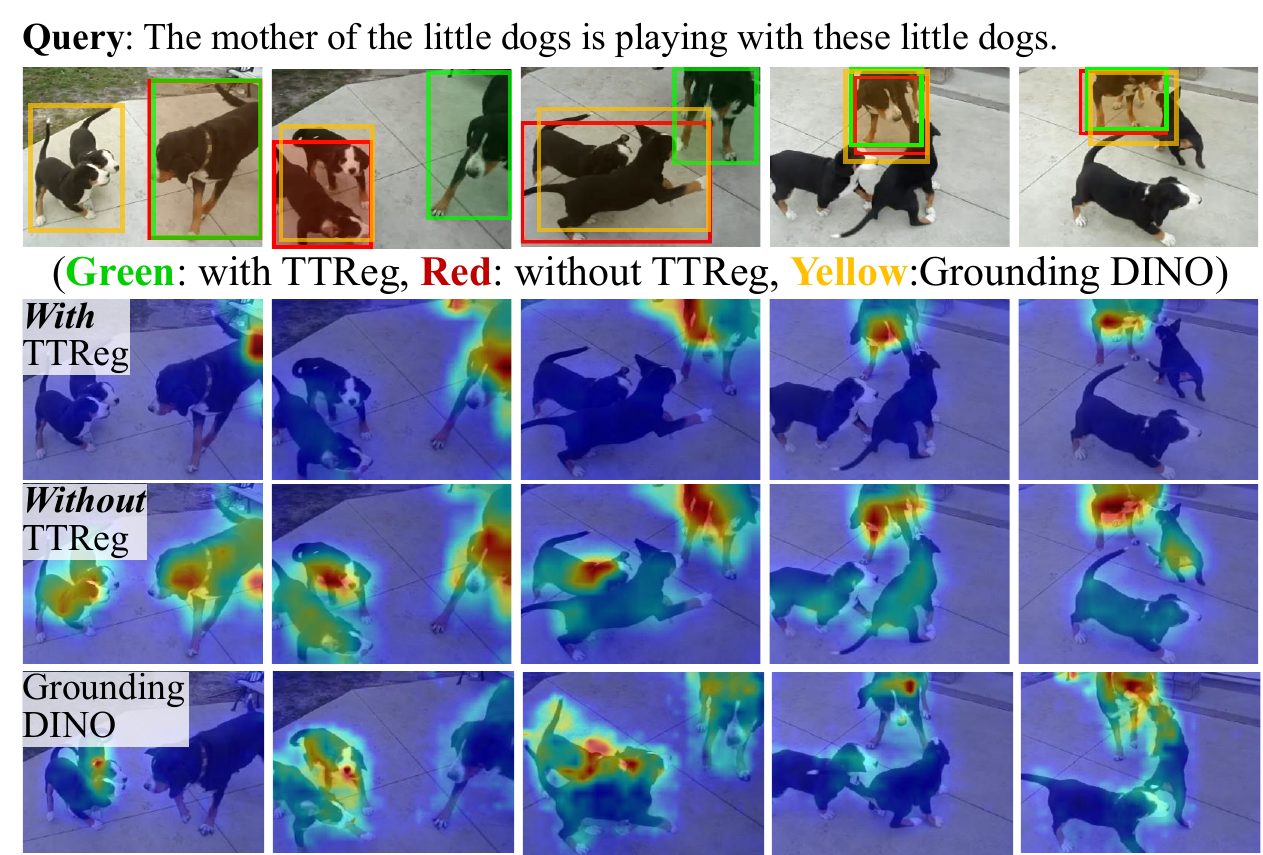}
\caption{\textbf{Attention and detection comparison} between the \texttt{[BOX]}-induced RST/text feature and image features (\textcolor{green}{green}: w/ TTReg; \textcolor{red}{red}: w/o TTReg). TTReg keeps attention and boxes on the target, while removing it causes scattered attention and spatial jitter. Grounding DINO (\textcolor{orange}{yellow} boxes) instead uses text–image attention that focuses on a distractor.}
\label{fig:attn_ttreg}
\end{figure}

\subsection{Temporal Consistency Regularization}
\label{sec:tce}
Although the OVD localizes all frames in parallel,
its outputs are still frame-wise predictions.
To obtain a temporally consistent trajectory, we further associate and regularize the query outputs across time.
A straightforward solution \cite{liang2025referdino} is to match per-frame queries with the \textit{Hungarian} algorithm.
However, due to appearance variation and query drift,
such association is often unstable.
To address this,
we first associate detector queries across frames to form candidate tubes,
and then apply a Tube-Mined Temporal Regularization (TTReg) to mine the best-aligned tube and encourage smooth and reliable trajectories during training.


\subsubsection{Memory-based Tube Association.}
From the frame-wise detector outputs $\{(\mathbf{q}_t^i, b_t^i)\}_{i=1}^{N}$,
we first keep the top-$N_q$ queries in each frame, denoted as $\{(\mathbf{q}_t^i, b_t^i)\}_{i=1}^{N_q}$.
These queries are then associated across time to form candidate tubes.
For the first frame $t=1$, we initialize a reference memory:
\begin{equation}
    \mathbf{M}_1 = \{\hat{\mathbf{q}}_1^i\}_{i=1}^{N_q},
\end{equation}
where $\hat{\mathbf{q}}_1^i = \mathbf{q}_1^i$.
For each subsequent frame $t>1$,
we match its queries $\{\mathbf{q}_t^i\}_{i=1}^{N_q}$ to the memory from the previous frame $\mathbf{M}_{t-1}$ using the \textit{Hungarian} algorithm based on cosine similarity.
The matched queries are re-ordered to align with the memory slots,
and the memory is updated by an exponential moving average:
\begin{equation}
    \mathbf{M}_t = (1-\alpha_t)\mathbf{M}_{t-1} + \alpha_t \,\mathtt{reorder}(\{\mathbf{q}_t^i\}_{i=1}^{N_q}),
    \label{eq:memory_update}
\end{equation}
where $\alpha_t$ is an adaptive update rate.
In this way,
each memory slot is encouraged to follow the same object instance over time,
producing $N_q$ candidate tubes:
\begin{equation}
    \Big\{\{(\hat{\mathbf{q}}_t^i,\hat{b}_t^i)\}_{t=1}^{T}\Big\}_{i=1}^{N_q}.
\end{equation}
This association is still imperfect in practice,
since the detector queries may drift when the target changes appearance across frames.
As illustrated in Fig. \ref{fig:attn_ttreg},
the attention inside the object region fluctuates across frames,
yielding unstable similarity between the corresponding per-frame queries.
We therefore further regularize the associated tubes via TTReg during training, as mentioned below.

\subsubsection{Tube-mined Temporal Regularization}
TTReg consists of two steps.
It first mines the tube that best aligns with the ground truth from the associated candidate tubes, and then supervise its feature and geometric consistency across frames.

\noindent
\textbf{Ground-Truth-Aligned Tube Mining.}
Among the $N_q$ candidate tubes, we select the one that best matches the ground truth and use it as the target trajectory for temporal supervision. Following Grounding-DINO~\cite{liu2024grounding}, we compute three spatial-semantic costs for each candidate tube: a classification cost $C_{\text{cls}}$, a box regression cost $C_{\text{bbox}}$, and a GIoU cost $C_{\text{giou}}$.
To further favor temporally smooth predictions,
we introduce a temporal consistency cost defined as the average ``1 - GIoU'' over all adjacent frame pairs:
\begin{equation}
    C_{\text{temp}}^i =
    \frac{1}{T-1}\sum\nolimits_{t=1}^{T-1}
    \left(1-\mathrm{GIoU}(\hat{b}_t^i,\hat{b}_{t+1}^i)\right),
    \label{eq:temporal_cost}
\end{equation}
where $\hat{b}_t^i$ is the predicted box of the $i$-th candidate tube at frame $t$. The final matching cost is
\begin{equation}
    C^i =
    \lambda_{\text{cls}} C_{\text{cls}}^i +
    \lambda_{\text{bbox}} C_{\text{bbox}}^i +
    \lambda_{\text{giou}} C_{\text{giou}}^i +
    \lambda_{\text{temp}} C_{\text{temp}}^i.
\end{equation}
This encourages the model to select a predicted tube that exhibits high spatial coherence.
In this way, the candidate tube with the lowest cost is selected as the best-aligned tube:
\begin{equation}
    \{(\mathbf{q}_t^*, b_t^*)\}_{t=1}^{T}.
\end{equation}


\noindent
\textbf{Cross-Frame Temporal Regularization.}
After identifying the best-aligned tube, we explicitly regularize its feature and geometric consistency across adjacent frames. Since all frames are processed together, this supervision can be applied to the whole trajectory within a single training pass.
We use two losses:
\begin{align}
    \mathcal{L}_{\text{feat}} &=
    \frac{1}{T-1}\sum\nolimits_{t=1}^{T-1}
    \left(1-\frac{\mathbf{q}_t^* \cdot \mathbf{q}_{t+1}^*}
    {\|\mathbf{q}_t^*\|\,\|\mathbf{q}_{t+1}^*\|}\right), \nonumber \\
    \mathcal{L}_{\text{geom}} &=
    \frac{1}{T-1}\sum\nolimits_{t=1}^{T-1}
    \left(1-\mathrm{GIoU}(b_t^*, b_{t+1}^*)\right).
    \label{eq:loss}
\end{align}
Here, $\mathcal{L}_{\text{feat}}$ encourages stable query representations over time, while $\mathcal{L}_{\text{geom}}$ encourages smooth box trajectories.
By jointly optimizing these two losses,
the detector learns to produce temporally coherent queries and boxes,
turning frame-wise localization into robust tube-level grounding.


\begin{table*}[!t]
\centering
\small
\caption{\textbf{Performance comparison on STVG benchmarks.} We evaluate our method on HC-STVG v1/v2~\cite{STGVT} and VidSTG~\cite{zhang2020does} (including both declarative and interrogative queries). Evaluation metrics include mean temporal IoU (m\_tIoU), mean spatio-temporal IoU (m\_vIoU), and vIoU@$\tau$ (the fraction of predictions with a vIoU $\ge \tau$). Best results are highlighted in \textbf{bold}.} 
\begin{tabular*}{\textwidth}{@{\extracolsep{\fill}} l c c c c c c c c c @{}}
\toprule
\multirow{2.5}{*}{\textbf{Model}} & \multirow{2.5}{*}{\textbf{Setting}} 
& \multicolumn{4}{c}{\textbf{HC-STVG v1}} 
& \multicolumn{4}{c}{\textbf{HC-STVG v2}} \\
\cmidrule(lr){3-6} \cmidrule(lr){7-10}
& & \textbf{m\_tIoU} & \textbf{m\_vIoU} & \textbf{vIoU@0.3} & \textbf{vIoU@0.5}
  & \textbf{m\_tIoU} & \textbf{m\_vIoU} & \textbf{vIoU@0.3} & \textbf{vIoU@0.5} \\
\midrule
STVGBert \cite{su2021stvgbert}     & \multirow{7}{*}{Fully Sup}
& - & 20.4  & 29.4  & 11.3
& -& -  & -  & - \\
TubeDETR \cite{yang2022tubedetr}   & 
& 43.7 & 32.4  & 49.8  & 23.5
& 53.9 & 36.4  & 58.8  & 30.6 \\
STCAT \cite{jin2022embracing}      &
& 49.4 & 35.1  & 57.7  & 30.1
& - & -  & -  & - \\
STVGFormer \cite{lin2023collaborative} & 
& - & 36.9  & 62.2  & 34.8
& 58.1 & 38.7  & 65.5  & 33.8 \\
VG-DINO \cite{wasim2024videogrounding} &
& - & 38.3  & 62.5  & 36.1
& - & 39.9  & 67.1  & 34.5 \\
CG-STVG \cite{gu2024context}       &
& 52.8 & 38.4  & 61.5  & 36.3
& 60.0 & 39.5  & 64.5  & 36.3 \\
TA-STVG \cite{gu2025knowing}       & 
& 53.0 & 39.1  & 63.1  & 36.8
& 60.4 & 40.2  & 65.8  & 36.7 \\
\midrule
Qwen3-VL-4B \cite{bai2025qwen3}       &
& 44.6 & 19.5 & 25.4 & 4.9
& 45.2 & 19.4 & 24.7 & 5.5 \\
Qwen3-VL-8B \cite{bai2025qwen3}       & \multirow{3}{*}{MLLM}
& 47.6 & 21.5 & 30.3 & 6.5
& 53.1 & 21.9 & 30.0 & 6.6 \\
SpaceVLLM \cite{spacevllm}            & 
& 56.9 & 39.3 & 66.6 & 36.9
& 58.0 & 34.0 & 56.9 & 24.7 \\
STVG-R1 \cite{zhang2026stvg}             & 
& 56.9 & 39.1 & 66.7 & 38.6
& 61.3 & 40.8 & 67.9 & 38.3 \\
\textbf{DEViL (Ours)}              & 
& \textbf{59.0} & \textbf{43.1} & \textbf{70.5} & \textbf{44.3}
& \textbf{61.7}    & \textbf{42.5} & \textbf{67.3} & \textbf{42.2} \\
\midrule
\midrule
\multirow{2.5}{*}{\textbf{Model}} & \multirow{2.5}{*}{\textbf{Setting}} 
& \multicolumn{4}{c}{\textbf{VidSTG (Declarative Sentence)}} 
& \multicolumn{4}{c}{\textbf{VidSTG (Interrogative Sentence)}} \\
\cmidrule(lr){3-6} \cmidrule(lr){7-10}
& & \textbf{m\_tIoU} & \textbf{m\_vIoU} & \textbf{vIoU@0.3} & \textbf{vIoU@0.5}
  & \textbf{m\_tIoU} & \textbf{m\_vIoU} & \textbf{vIoU@0.3} & \textbf{vIoU@0.5} \\
\midrule
STGVT \cite{STGVT}                 & \multirow{8}{*}{Fully Sup}
& - & 21.6  & 29.8  & 18.9
& -& -  & -  & - \\
STVGBert \cite{su2021stvgbert}     & 
& - & 24.0  & 30.9  & 18.4
& -& 22.5  & 26.0  & 16.0 \\
TubeDETR \cite{yang2022tubedetr}   & 
& 48.1 & 30.4  & 42.5  & 28.2
& 46.9 & 25.7  & 35.7  & 23.2 \\
STCAT \cite{jin2022embracing}      & 
& 50.8 & 33.1  & 46.2  & 32.6
& 49.7 & 28.2  & 39.2  & 26.6 \\
STVGFormer \cite{lin2023collaborative} & 
& - & 33.7  & 47.2  & 32.8
& - & 28.5  & 39.9  & 26.2 \\
CG-STVG \cite{gu2024context}       & 
& 51.4 & 34.0 & 47.7 & 33.1
& 49.9 & 29.0 & 40.5 & 27.5 \\
VG-DINO \cite{wasim2024videogrounding} &
& 52.0 & 34.7  & 48.1  & 34.0
& 50.8 & 29.9  & 41.0  & 27.6 \\
TA-STVG \cite{gu2025knowing}       &
& 51.7 & 34.4 & 48.2 & 33.5
& 50.2 & 29.5 & 41.5 & 28.0 \\
\midrule
Qwen3-VL-4B \cite{bai2025qwen3}       &
& 36.2 & 13.1 & 16.6 & 7.0
& 36.1 & 8.9 & 10.2 & 3.8 \\
Qwen3-VL-8B \cite{bai2025qwen3}       &
& 37.0 & 13.4 & 16.5 & 7.1
& 35.0 & 9.3 & 11.0 & 3.9 \\
LLaVA-ST \cite{llavast}            & \multirow{1}{*}{MLLM}
& 45.1 & 14.3 & 18.3 & 7.4
& 43.0 & 11.4 & 13.9 & 5.8 \\
SpaceVLLM \cite{spacevllm}            & 
& 47.7 & 27.4 & 39.1 & 26.2
& 48.5 & 25.4 & 35.9 & 22.2 \\
\textbf{DEViL (Ours)}              & 
& \textbf{50.2} & \textbf{33.6} & \textbf{46.5} & \textbf{34.0}
& \textbf{48.5} & \textbf{28.8} & \textbf{38.7} & \textbf{28.2} \\
\bottomrule
\end{tabular*}
\label{tab:stvg_results}
\end{table*}

\subsection{Progressive Optimization and Inference}
\label{sec:training}
To jointly fine-tune the MLLM with OVD in an end-to end way, we employ a progressive training strategy.
The overall architecture remains unchanged throughout training,
and no stage-specific modules or objectives are introduced.
The only difference across stages is the type of data used.
Specifically, the three stages serve three intuitive purposes:
(i) establishing the semantic bridge between the MLLM and the detector,
(ii) teaching temporal grounding to the MLLM,
and (iii) finally enabling full spatio-temporal collaboration.The corresponding training datasets and reformulated annotations will be publicly released.

\noindent
\textbf{Stage 1: \textit{Bridging} MLLM and OVD.}
The preliminary training stage focuses on establishing the fundamental connection between the MLLM and the detector.
Using image-based referring expression datasets (\textbf{\textit{RefCOCO}} \cite{refcoco}, \textbf{\textit{RefCOCO+}} \cite{refcoco}, \textbf{\textit{RefCOCOg}} \cite{refcocog}),
DEViL learns to adaptively output a task-specific \texttt{[BOX]} token that encapsulates referential semantics according to the input text.

\noindent
\textbf{Stage 2: Temporal \textit{Alignment} for MLLM.}
In this stage, we enhance the temporal alignment capability of the MLLM.
We freeze the OVD and fine-tune only the MLLM on temporal grounding datasets
(\textbf{\textit{TACoS}} \cite{tacos}, \textbf{\textit{ActivityNet Captions}} \cite{activitynet}, \textbf{\textit{QVHighlights}} \cite{qvhighlight}).
Instead of predicting bounding boxes,
the MLLM is required to determine when the referenced event occurs.

\noindent
\textbf{Stage 3: Spatio-Temporal \textit{Collaboration} Training.}
In the final stage, we unfreeze the whole architecture and jointly optimize the MLLM and the detector on a unified spatio-temporal corpus constructed from public datasets via data reformulation and pseudo-labeling.
Following LLaVA-ST~\cite{llavast}, we enhance and revise the textual queries in training data of \textbf{\textit{HC-STVG v1/v2}} \cite{STGVT} and \textbf{\textit{VidSTG}} \cite{zhang2020does} into instruction-style inputs suitable for MLLM training, while preserving their human-labeled spatio-temporal tubes.
To enrich spatial supervision, we re-purpose the static grounding set \textbf{SA-V}~\cite{sam2} by assigning time spans to obtain tube-level labels.
In addition, we implement an automatic pipeline that combines a strong detector (MM-Grounding-DINO \cite{mmgroundingdino}), a powerful REC VLM (VLM-R1 \cite{vlmr1}) and a tracker (SUTrack \cite{sutrack}) to generate dense object tubes,
which is then used to lift the Stage-2 temporal grounding datasets (\textbf{\textit{TACoS}}, \textbf{\textit{ActivityNet Captions}}, \textbf{\textit{QVHighlights}}) to the spatio-temporal setting.
The final corpus contains 196k spatio-temporally annotated samples, combining human-labeled and automatically generated tubes.
During this stage, we apply Tube-mined Temporal Regularization (TTReg) to enforce cross-frame alignment within tube-level queries: the MLLM produces RSTs and the detector executes them, forming a collaborative closed loop between high-level reasoning and low-level grounding.
To foster reproducibility, we will release reformatted annotations, automatic-labeling method and pseudo labels.

\noindent
\textbf{Unified Inference for Multiple Tasks.}
At inference, DEViL runs in an \emph{intent-conditioned} manner: the MLLM either returns text-only outputs or emits the special \texttt{[BOX]} token, its associated RST, and the predicted temporal interval $[t_{\text{start}}, t_{\text{end}}]$ to trigger the OVD for spatial grounding.
For grounding-required cases, the OVD together with the memory-based tube association in Eq.~\eqref{eq:memory_update} produces $N_q$ tube hypotheses
$\{\tau_i\}_{i=1}^{N_q}$, where
$\tau_i = \{(s_t^i, b_t^i)\}_{t=1}^{T}$ denotes the confidence $s_t^i$ given by the classification head of OVD and box $b_t^i$ of the $i$-th query at frame $t$.
The final tube index $\hat{i}$ is selected by calculating the average confidence strictly within the temporal segment predicted by the MLLM:
\begin{equation}
\hat{i} = \arg\max_{i \in [1,N_q]} \frac{1}{t_{\text{end}} - t_{\text{start}}}\sum\nolimits_{t=t_{\text{start}}}^{t_{\text{end}}} s_t^i .
\end{equation}
We then output the corresponding trajectory $\{b_t^{\hat{i}}\}_{t=t_{\text{start}}}^{t_{\text{end}}}$ as the final spatio-temporal grounding evidence.

\label{sec:ts}

\begin{table}[t]
\centering
\small
\setlength{\tabcolsep}{5pt}
\caption{Results on the NExT-GQA \cite{xiao2024can} grounded VideoQA benchmark.
Acc@GQA measures grounded QA accuracy, while mIoP/IoP@0.5 and mIoU/IoU@0.5 evaluate temporal localization and spatial overlap between predicted and ground-truth evidence segments. Bold numbers denote the best performance.}
\resizebox{\columnwidth}{!}{
\begin{tabular}{@{} l c c c c c @{}}
\toprule
\textbf{Model} & \textbf{Acc@GQA} & \textbf{mIoP} & \textbf{IoP@0.5} & \textbf{mIoU} & \textbf{IoU@0.5} \\
\midrule
VIOLETv2 \cite{fu2021violet}           & 12.8 & 23.6 & 23.3 & 3.1  & 1.3 \\
SeViLA \cite{yu2023self}             & 16.6 & 29.5 & 22.9 & 21.7 & 13.8 \\
LangRepo \cite{kahatapitiya2025language}           & 17.1 & 31.3 & 28.7 & 18.5 & 12.2 \\
FrozenBiLM NG+ \cite{yang2022zero}     & 17.5 & 24.2 & 23.7 & 9.6  & 6.1 \\
VideoStreaming \cite{qian2024streaming}     & 17.8 & 32.2 & 31.0 & 19.3 & 13.3 \\
LLoVi \cite{zhang2024simple}              & 24.3 & 37.3 & 36.9 & 20.0 & 15.3 \\
Grounded-VideoLLM \cite{wang2024grounded}   & 26.7 & 34.5 & 34.4 & 21.1 & 18.0 \\
HawkEye \cite{wang2024hawkeye}            & -- & -- & -- & 25.7 & 19.5 \\
VideoChat-TPO \cite{yan2025task}                              & 25.5 & 35.6 & 32.8 & 27.7 & 23.4 \\
\textbf{DEViL (Ours)}    & \textbf{37.1} & \textbf{49.1} & \textbf{49.6} & \textbf{28.9} & \textbf{25.0} \\
\bottomrule
\end{tabular}
}
\label{tab:nextgqa_results_half}
\end{table}

\begin{table}[t]
\centering
\small
\setlength{\tabcolsep}{4pt}
\caption{Performance on the Charades-STA \cite{Charades} temporal grounding benchmark. Bold numbers denote the best performance. The methods marked in gray$^*$ represent fine-tuning on corresponding benchmarks, while those in black indicate zero-shot settings.}
\resizebox{\columnwidth}{!}{
\begin{tabular}{@{} l c c c c @{}}
\toprule
\multirow{2.5}{*}{\textbf{Model}} & \multicolumn{4}{c}{\textbf{Charades-STA}} \\
\cmidrule(lr){2-5}
& \textbf{R1@0.3} & \textbf{R1@0.5} & \textbf{R1@0.7} & \textbf{m\_tIoU} \\
\midrule
Video-LLaMA\cite{zhang2023video}             & 25.2 & 10.6 & 3.4  & 16.8 \\
Video-ChatGPT\cite{maaz2024video}          & 27.2 & 6.2  & 1.9  & 19.7 \\
VideoChat\cite{2023videochat}              & 32.8 & 8.6  & 0.0  & 25.9 \\
Momenter\cite{qian2024momentor}               & 42.6 & 26.6 & 11.6 & 28.5 \\
VTimeLLM\cite{huang2024vtimellm}               & 51.0 & 27.5 & 11.4 & 31.2 \\
TimeChat\cite{ren2024timechat}               & --   & 32.2 & 13.4 & --   \\
VTG-LLM\cite{guo2025vtg}                & --   & 33.8 & 15.7 & --   \\
HawkEye\cite{wang2024hawkeye}                & 50.6 & 31.4 & 14.5 & 33.7 \\
Grounded-VideoLLM\cite{wang2024grounded}      & 54.2 & 36.4 & 19.7 & 36.8 \\
TRACE                               & - & 40.3 & 19.4 & - \\
LLaVA-ST\cite{llavast}               & 63.1 & 44.8 & 23.4 & 42.4 \\
TimeSuite                            & 69.9 & 48.7 & 24.0 & - \\
\rowcolor{gray!15} HawkEye*\cite{wang2024hawkeye}                & 72.5 & 58.3 & 28.8 & 49.3 \\
\rowcolor{gray!15} TimeSuite* & 79.4 & 67.1 & 43.0 & - \\

\textbf{DEViL}         & \textbf{72.6} & \textbf{51.5} & \textbf{25.2} & \textbf{47.7} \\
\rowcolor{gray!15} \textbf{DEViL*}         & 79.8 & 66.5 & 42.8 & 58.5 \\
\bottomrule
\end{tabular}
}
\label{tab:charades_results}
\end{table}

\section{Experiments}
\label{sec:experiments}

In this section, we first elaborate on the implementation details (Sec. \ref{sec:detail}) of DEViL.
Next, we conduct extensive testing (Sec. \ref{sec:compare}) under \textit{spatio-temporal video grounding}, \textit{temporal video grounding}, \textit{grounded video question and answering}, and \textit{common video understanding} to verify its effectiveness and generalization.
Finally, we validate the effectiveness of each module by ablation studies (Sec. \ref{sec:ablation}).

\subsection{Implementation Details}
\label{sec:detail}
We initialize the vision encoder with SigLIP~\cite{zhai2023sigmoid} and the LLM with Qwen2.5-7B~\cite{yang2025qwen3}; checkpoints are loaded from the public VideoLLaMA3-7B~\cite{zhang2025videollama} release.
The projector is a two-layer MLP with GELU.
The OVD is Grounding DINO~\cite{liu2024grounding} with a Swin-B~\cite{liu2021swin} backbone.
Across Stage-1/2/3 we freeze the MLLM vision encoder and fine-tune the LLM with LoRA~\cite{hu2022lora} ($\alpha=512$, $r=256$).
In Stage-2 the detector is frozen, while in Stages-1 and -3 it is tunable.
We use AdamW for optimization.
The detector and the LLM use a learning rate of $1\times10^{-4}$ in all trainable stages, while the projector uses $1\times10^{-5}$.
Clips are uniformly sampled to $T=64$ frames per video.
We set the EMA update rate to $\alpha_t=0.1$.
In the Hungarian matcher and TTReg, following DETR~\cite{detr}, the cost weights are set to $\lambda_{\text{cls}}=1$, $\lambda_{\text{bbox}}=5$, $\lambda_{\text{giou}}=3$, $\lambda_{\text{feat}}=2$, and $\lambda_{\text{geom}}=1$., which are then used in training.
All experiments are conducted on 8$\times$A800 GPUs.

\subsection{Main Comparisons}
\label{sec:compare}

\noindent
\textbf{Spatio-Temporal Video Grounding.} We evaluate \textbf{DEViL} on two primary STVG benchmarks: HC-STVG \cite{STGVT} and VidSTG \cite{zhang2020does}, with quantitative results summarized in Table~\ref{tab:stvg_results}.

\textbf{HC-STVG Datasets.} The HC-STVG benchmark (including v1 and v2) focuses on human-centric scenarios, requiring models to spatio-temporally localize complex and continuous human actions.
As shown in Table~\ref{tab:stvg_results}, DEViL achieves state-of-the-art performance across all metrics on both versions. 
Specifically, it attains an impressive m\_vIoU of 43.1 on v1 and 42.5 on v2, outperforming not only recent MLLM-based methods (e.g., SpaceVLLM \cite{spacevllm} and STVG-R1 \cite{zhang2026stvg}) but also strong fully-supervised algorithms (e.g., TA-STVG \cite{gu2025knowing}). 
This demonstrates that DEViL's detector-empowered architecture and temporal regularization effectively produce accurate and temporally consistent bounding boxes for complex human movements.

\textbf{VidSTG Dataset.} The VidSTG benchmark provides a broader range of object categories and relations. Crucially, it evaluates models using two distinct types of language queries: declarative sentences (descriptions) and interrogative sentences (questions), thoroughly testing the model's reasoning and grounding flexibility. On this benchmark, DEViL exhibits strong robustness across different query styles. For declarative sentences, it achieves a 50.2 m\_tIoU and 33.6 m\_vIoU, significantly surpassing previous MLLMs such as LLaVA-ST \cite{llavast} and SpaceVLLM \cite{spacevllm}. More importantly, on the challenging interrogative queries that require deeper reasoning, DEViL maintains a highly competitive 28.8 m\_vIoU, dominating other MLLM baselines and rivaling specialized fully-supervised systems. These results suggest that coupling an OVD with an MLLM through the proposed RST enables more reliable fine-grained visual grounding under diverse query styles, while narrowing the gap between general video understanding and spatio-temporal grounding.

\noindent
\textbf{Temporal Video Grounding (TVG).}
As shown in Table~\ref{tab:charades_results}, we evaluate \textbf{DEViL} on the Charades-STA benchmark. In the zero-shot setting, it achieves 51.5 R1@0.5 and 47.7 mean temporal IoU, surpassing Grounded-VideoLLM \cite{wang2024grounded} and improving over LLaVA-ST \cite{llavast} by 6.7 R1@0.5 and 5.3 m\_tIoU. When fine-tuned (\textbf{DEViL*}), performance further scales to 66.5 R1@0.5 and 58.5 m\_tIoU. These results confirm DEViL's strong temporal localization capabilities among current video MLLMs.

\noindent
\textbf{Grounded Video QA (GQA).}
As shown in Table~\ref{tab:nextgqa_results_half}, we evaluate \textbf{DEViL} on the NExT-GQA benchmark, which jointly measures answering accuracy and evidence grounding. DEViL achieves 37.1 Acc@GQA, 49.1 mIoP, and 28.9 mIoU, clearly surpassing recent video grounding QA models such as LLoVi \cite{zhang2024simple}, Grounded-VideoLLM \cite{wang2024grounded}, HawkEye \cite{wang2024hawkeye}, and VideoChat-TPO \cite{yan2025task}.

\begin{figure*}[t]
\centering
\includegraphics[width=1\linewidth]{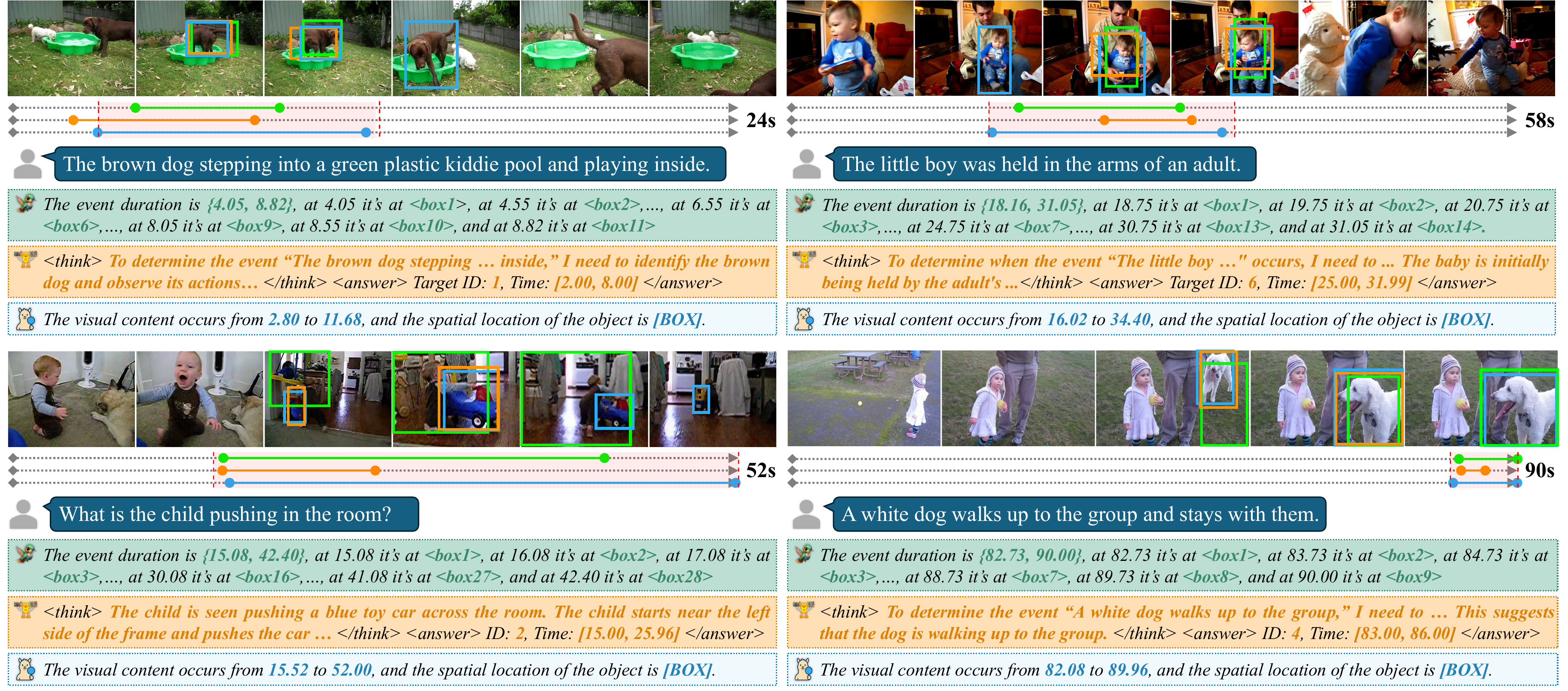}
\caption{\textbf{Qualitative comparison among LLaVA-ST, STVG-R1, and DEViL.} 
For each example, the predictions of LLaVA-ST, STVG-R1, and \textbf{DEViL} are shown in \textcolor{green}{green}, \textcolor{orange}{yellow}, and \textcolor{blue}{blue}, respectively. The light red shaded region on the timeline represents the ground-truth (GT) temporal interval.}
\label{fig:compare}
\end{figure*}

\noindent
\textbf{Video Understanding.}
As shown in Table~\ref{tab:mvbench_tempcompass_videomme_half}, we further assess \textbf{DEViL} on generic video understanding benchmarks, including MVBench \cite{li2024mvbench}, TempCompass \cite{liu2024tempcompass}, and VideoMME \cite{fu2025video}. DEViL attains 65.6 on MVBench, close to the best score of TimeMarker \cite{chen2024timemarker}, and achieves the highest accuracy on TempCompass with 63.75. On VideoMME, it also sets new highs with 58.7 without subtitles and 63.3 with subtitles, outperforming prior video LLMs such as TimeChat \cite{ren2024timechat}, VideoLLaMA2 \cite{cheng2024videollama}, VideoChat2 \cite{li2024mvbench}, Video-LLaVA \cite{lin2024video}, LLaVA-NeXT-Video \cite{li2024llava}, and LLaVA-ST \cite{li2025llava}, indicating strong general video reasoning ability.

\begin{table}[t]
\centering
\small
\setlength{\tabcolsep}{5pt}
\caption{Comparison on MVBench \cite{li2024mvbench}, TempCompass \cite{liu2024tempcompass}, and VideoMME \cite{fu2025video} (without/with subtitles); all numbers are average accuracy scores. Bold numbers denote the best performance.}
\resizebox{\columnwidth}{!}{
\begin{tabular}{@{} l c c c @{}}
\toprule
\multirow{2}{*}{\textbf{Model}} & \multirow{2}{*}{\textbf{MVBench}} & \multirow{2}{*}{\textbf{TempCompass}}& \textbf{VideoMME} \\
&&& (\textit{w/o} / \textit{w/} subs)\\
\midrule
TimeChat-7B \cite{ren2024timechat}            & 38.5 & 38.9  & 34.7 / -- \\
VideoLLaMA2-7B \cite{cheng2024videollama}         & 54.6 & 43.4  & 47.9 / -- \\
VideoChat2-7B \cite{li2024mvbench}          & 51.1 & 48.81 & 42.3 / 54.6 \\
Video-LLaVA-7B \cite{lin2024video}         & 43.0 & 49.77 & 39.9 / 41.6 \\
LLaVA-NeXT-Video-7B \cite{li2024llava}    & 53.1 & 53.75 & 33.7 / -- \\
TimeMarker \cite{chen2024timemarker}             & \textbf{67.4} & 60.4  & 57.3 / -- \\
LLaVA-ST \cite{li2025llava}               & 64.2 & --    & --   / -- \\
\textbf{DEViL (Ours)}          & 65.6 & \textbf{63.75} & \textbf{58.7 / 63.3} \\
\bottomrule
\end{tabular}
}
\label{tab:mvbench_tempcompass_videomme_half}
\end{table}

\noindent
\textbf{Qualitative analysis.}
Fig. \ref{fig:compare} qualitatively compares LLaVA-ST, STVG-R1, and DEViL on self-collected zero-shot video samples. While all models comprehend the text queries, their distinct spatial mechanisms lead to different failure modes. LLaVA-ST relies on autoregressive coordinate decoding, which is prone to accumulated localization errors and often leads to unstable spatial trajectories and inaccurate temporal boundaries. Conversely, STVG-R1 employs a disjointed ``segment-then-select'' paradigm (prompting the LLM to choose pre-extracted SAM region IDs), which bottlenecks spatial precision on upstream proposals and lacks cross-frame temporal coherence. In contrast, by directly coupling the MLLM with an open-vocabulary detector via RST and temporal regularization, DEViL bypasses both sequence explosion and rigid proposals, yielding highly accurate temporal intervals and stable bounding box trajectories.

\subsection{Ablation Study}
\label{sec:ablation}
We next dissect the core components of DEViL to understand their individual contributions to the overall performance. We systematically ablate the RST (denoted as the \texttt{[BOX]} token) and the impact of training data scale, followed by the proposed TTReg and the test-time MTA. Finally, we benchmark the inference efficiency of our architecture. Unless otherwise stated, ablations are conducted following the settings in Sec.~\ref{sec:experiments}.

\noindent
\textbf{RST and Data Scaling.}
Table~\ref{tab:vidstg_ablation} ablates the \texttt{[BOX]} token (RST) and supervised fine-tuning (SFT) scale on VidSTG. A baseline replacing RST with raw text embeddings (w/o \texttt{[BOX]}) drastically degrades spatial grounding (e.g., m\_vIoU drops to 20.4 on declarative queries). This gap verifies RST's superiority: it distills reasoning context (\textit{Deep Semantics}) and serves as a learnable localization trigger (\textit{Dynamic Instruction}). Furthermore, DEViL trained on a 101K SFT subset (excluding VidSTG/HC-STVG) exhibits strong \textbf{zero-shot} grounding capabilities. Scaling to the full 196K corpus ultimately yields the best spatio-temporal performance.

\noindent
\textbf{TTReg: GTM and CFR.}
Table~\ref{tab:ablation_tmtr} dissects TTReg on HC-STVG v1/v2. Compared to the baseline, enabling Ground-Truth-Aligned Tube Mining (\textbf{GTM}) alone improves both spatial and temporal metrics by selecting tubes via temporal cost. Activating Cross-Frame Temporal Regularization (\textbf{CFR}) alone explicitly penalizes feature and geometric inconsistencies, significantly boosting m\_tIoU. Combining both (\textbf{GTM+CFR}) yields the best synergy (59.0/43.1 on v1), proving that GTM mines cleaner tubes while CFR effectively stabilizes them across frames.

\begin{table}[t]
\centering
\small
\setlength{\tabcolsep}{4pt}
\caption{\textbf{Ablation study of TTReg on HC-STVG v1/v2.} We evaluate the model by toggling Ground-Truth-Aligned Tube Mining (GTM) and Cross-Frame Temporal Regularization (CFR).}
\resizebox{\columnwidth}{!}{
\begin{tabular}{@{} cc cc cc @{}}
\toprule
\multicolumn{2}{c}{\textbf{TTReg Modules}} & \multicolumn{2}{c}{\textbf{HC-STVG v1}} & \multicolumn{2}{c}{\textbf{HC-STVG v2}} \\
\cmidrule(lr){1-2}\cmidrule(lr){3-4}\cmidrule(lr){5-6}
\textbf{GTM} & \textbf{CFR} & \textbf{m\_tIoU} & \textbf{m\_vIoU} & \textbf{m\_tIoU} & \textbf{m\_vIoU} \\
\midrule
\xmark & \xmark & 56.8 & 41.5 & 60.2 & 41.0 \\
\cmark & \xmark & 58.1 & 42.2 & 60.8 & 41.6 \\
\xmark & \cmark & \textbf{59.3} & 41.9 & 61.3 & 40.6 \\
\cmark & \cmark & 59.0 & \textbf{43.1} & \textbf{61.7} & \textbf{42.5} \\
\bottomrule
\end{tabular}
}
\label{tab:ablation_tmtr}
\end{table}

\noindent
\textbf{Memory-based Tube Association.}
Table~\ref{tab:ablation_mta} isolates test-time MTA. Without TTReg, MTA brings marginal spatial refinement. However, when coupled with a TTReg-trained model, MTA yields a stronger synergistic boost (+0.7 m\_vIoU on v1) without compromising m\_tIoU. Thus, MTA is an efficient test-time plug-in whose spatial tracking benefits are distinctly amplified by temporally regularized training.

\noindent
\textbf{Efficiency Analysis.}
To validate our solution to the sequence explosion problem of textualized decoding, Table~\ref{tab:efficiency_new} compares inference efficiency on a single A800 GPU. By processing the whole video in a single forward pass, DEViL achieves a leading \textbf{14.33} FPS—doubling LLaVA-ST and vastly outperforming STVG-R1. While its memory usage is slightly higher due to the parallel OVD-based spatial grounding over video frames, this overhead remains moderate (37.4 GB) and enables a clearly superior speed--accuracy trade-off, highlighting DEViL as a highly efficient paradigm for video grounding.

\begin{table}[t]
\centering
\small
\setlength{\tabcolsep}{4pt}
\caption{\textbf{Ablation study of Memory-based Tube Association (MTA).} Performance comparison on HC-STVG v1/v2 with and without test-time MTA across different TTReg training settings.}
\resizebox{\columnwidth}{!}{
\begin{tabular}{@{} cc cc cc @{}}
\toprule
\multirow{2.5}{*}{\textbf{TTReg (train)}} &
\multirow{2.5}{*}{\textbf{MTA (test)}} &
\multicolumn{2}{c}{\textbf{HC-STVG v1}} &
\multicolumn{2}{c}{\textbf{HC-STVG v2}} \\
\cmidrule(lr){3-4}\cmidrule(lr){5-6}
 &  & \textbf{m\_tIoU} & \textbf{m\_vIoU} & \textbf{m\_tIoU} & \textbf{m\_vIoU} \\
\midrule
\xmark & \xmark & 56.8 & 41.1 & 60.2 & 40.6 \\
\xmark & \cmark & 56.8 & 41.5 & 60.2 & 41.0 \\
\cmark & \xmark & 59.0 & 42.4 & 61.7 & 41.8 \\
\cmark & \cmark & \textbf{59.0} & \textbf{43.1} & \textbf{61.7} & \textbf{42.5} \\
\bottomrule
\end{tabular}
}
\label{tab:ablation_mta}
\end{table}

\begin{table}[t]
\centering
\small
\setlength{\tabcolsep}{4pt}
\caption{\textbf{Ablation study on VidSTG.} Comparison of configurations with and without the \texttt{[BOX]} token (RST) across different supervised fine-tuning (SFT) data scales.}
\begin{tabular}{@{} cc cc cc @{}}
\toprule
\multirow{2.5}{*}{\makecell{\textbf{[BOX]}\\\textbf{Token}}} & 
\multirow{2.5}{*}{\makecell{\textbf{SFT}\\\textbf{Data}}} & 
\multicolumn{2}{c}{\textbf{VidSTG Decla.}} & 
\multicolumn{2}{c}{\textbf{VidSTG Inter.}} \\
\cmidrule(lr){3-4}\cmidrule(lr){5-6}
& & \textbf{m\_tIoU} & \textbf{m\_vIoU} & \textbf{m\_tIoU} & \textbf{m\_vIoU} \\
\midrule
\xmark & 196K & 48.6 & 20.4 & 47.6 & 14.7 \\
\cmark & 101K & 44.4 & 25.2 & 43.2 & 18.2 \\
\cmark & 196K & \textbf{50.2} & \textbf{33.6} & \textbf{48.5} & \textbf{28.8} \\
\bottomrule
\end{tabular}
\label{tab:vidstg_ablation}
\end{table}

\begin{table}[t]
\centering
\small
\setlength{\tabcolsep}{6pt}
\caption{\textbf{Inference efficiency comparison.} Evaluation of model parameters, frames per second (FPS), and peak memory usage on a single A800 GPU with 64-frame inputs.}
\begin{tabular}{@{} l ccc @{}}
\toprule
\textbf{Models} & \textbf{Params} & \textbf{FPS}~$\uparrow$ & \textbf{PeakMem}~$\downarrow$ \\
\midrule
LLaVA-ST & 7B & 7.09 & 31.1 GB \\
Sa2VA    & 8B & 10.76 & 46.4 GB \\
STVG R1  & 7B & 1.23 & \textbf{28.6 GB} \\
\textbf{DEViL}  & 7B & \textbf{14.33} & 37.4 GB \\
\bottomrule
\end{tabular}
\label{tab:efficiency_new}
\end{table}

\section{Conclusion}
\label{sec:conclusion}


In this paper, we revisit MLLM-based spatio-temporal video grounding from an efficiency perspective. We show that existing methods, despite their different formulations, remain limited by either expensive autoregressive spatial decoding or heavy candidate construction on untrimmed videos. To address this issue, we propose DEViL, a detector-empowered Video-LLM framework that decomposes the task into sparse temporal grounding in the MLLM and dense spatial localization in a fully parallelizable detector. This design is efficient, minimally invasive to the MLLM, and able to directly leverage the strong spatial localization capability of a well-trained detector.
Built on this decomposition,
DEViL connects language reasoning and detector-based perception through a reference-semantic token,
and further improves temporal coherence with temporal consistency regularization.
Extensive experiments show that DEViL achieves a strong efficiency--accuracy trade-off and delivers outstanding performance on spatio-temporal video grounding.
We hope this work offers a practical step toward scalable,
evidence-grounded video-language systems for understanding and reasoning.

\bibliographystyle{ACM-Reference-Format}
\bibliography{main}

\clearpage
\def\DeViLIncludeSupplement{}
\newif\ifdevilsupplementstandalone
\ifdefined\DeViLIncludeSupplement
  \devilsupplementstandalonefalse
\else
  \devilsupplementstandalonetrue
\fi

\ifdevilsupplementstandalone
\documentclass[sigconf,authordraft]{acmart}
\AtBeginDocument{%
  \providecommand\BibTeX{{
    Bib\TeX}}}

\usepackage{amsmath}
\usepackage{amssymb}
\usepackage{multirow} 
\usepackage{tabularx}
\usepackage{marvosym}
\usepackage{colortbl}
\usepackage{pifont}
\usepackage{makecell}
\usepackage{booktabs}
\usepackage{balance}
\usepackage{algorithm}
\usepackage{algpseudocode}

\definecolor{mygray}{gray}{.9}
\definecolor{lightgray}{gray}{.95}
\newcommand{\red}[1]{{\color{red}#1}}
\newcommand{\green}[1]{{\color{green}#1}}
\newcommand{\blue}[1]{{\color{blue}#1}}
\newcommand{\todo}[1]{{\color{red}#1}}
\newcommand{\TODO}[1]{\textbf{\color{red}[TODO: #1]}}
\newcommand{\xf}[1]{{\color{black}#1}}
\setcopyright{none}
\settopmatter{
  printacmref=false,
  printccs=false,
  printfolios=false
}
\renewcommand\footnotetextcopyrightpermission[1]{}

\setcopyright{acmlicensed}
\copyrightyear{2018}
\acmYear{2018}
\acmDOI{XXXXXXX.XXXXXXX}
\acmConference[Conference acronym 'XX]{Make sure to enter the correct
  conference title from your rights confirmation email}{June 03--05,
  2018}{Woodstock, NY}
\acmISBN{978-1-4503-XXXX-X/2018/06}




\begin{document}

\title{Supplementary Material for ``Detector-Empowered Video Large Language Model for Efficient Spatio-Temporal Grounding''}

\maketitle
\fi

\appendix
\ifdevilsupplementstandalone\else
\section*{Supplementary Material}
\fi

\begin{table}[t]
\caption{\textbf{Comparison of HC-STVG v1/v2 \cite{STGVT}, VidSTG \cite{zhang2020does} and our self-collected data} in the Stage-3 spatio-temporal training corpus.
Samples denotes the number of tube-level training instances after instruction reformulation, with pseudo labeling applied where needed.}
\centering
\small
\setlength{\tabcolsep}{6pt}
\renewcommand{\arraystretch}{1.1}
\begin{tabular}{@{} l r r r @{}} 
\toprule
Dataset & \#Videos & \#Queries & \#Samples \\ 
\midrule
HC-STVG v1~\cite{STGVT}        & 4{,}500  & 4{,}500   & 4.5k  \\
HC-STVG v2~\cite{STGVT}        & 10{,}131 & 10{,}131  & 10k   \\
VidSTG~\cite{zhang2020does}    & 5{,}436  & 80{,}684  & 81k   \\
Self-collected (ours)          & \textbf{42{,}792} & \textbf{101{,}080} & \textbf{101k}  \\
\bottomrule
\end{tabular}

\label{tab:stvg_corpus_stats}
\end{table}


\section*{Overview}
For a better understanding of this work, we provide additional details, analysis, and results in this supplementary material as follows:
\begin{itemize}
    \item \textbf{Spatio-Temporal Training Data Details}: We describe the self-collected data, the auto-labeling pipeline, and the instruction-style reformulation used in Stage-3 training (Sec. \ref{sec:supp_training}).
    \item \textbf{Additional Analysis}: We provide an additional analysis on zero-shot generalization (Sec. \ref{sec:more experiments}).
    \item \textbf{Additional Qualitative Analysis}: We present more qualitative results across image and video grounding tasks to further illustrate the model's behavior (Sec. \ref{sec:visualized results}).
    \item \textbf{Limitations and Future Work Discussion}: We discuss the current limitation of single-target grounding and possible future extensions toward multi-target grounding (Sec. \ref{sec:limitations}).
\end{itemize}

\begin{table*}[t]
\caption{Spatio-temporal video grounding results on VidSTG under the declarative and interrogative settings. ``SFT Data'' denotes the amount of spatio-temporal supervision used during supervised fine-tuning. Rows marked with \textbf{*} indicate settings without VidSTG supervision.}
\centering
\normalsize
\setlength{\tabcolsep}{7pt}
\renewcommand{\arraystretch}{1.08}

\begin{tabular}{@{}cccccccccc@{}}
\toprule
\multirow{2}{*}{\textbf{Model}} 
& \multirow{2}{*}{\textbf{SFT Data}}
& \multicolumn{4}{c}{\textbf{VidSTG (Declarative)}} 
& \multicolumn{4}{c}{\textbf{VidSTG (Interrogative)}} \\
\cmidrule(lr){3-6} \cmidrule(lr){7-10}
& 
& \textbf{m\_tIoU} & \textbf{m\_vIoU} & \textbf{vIoU@0.3} & \textbf{vIoU@0.5}
& \textbf{m\_tIoU} & \textbf{m\_vIoU} & \textbf{vIoU@0.3} & \textbf{vIoU@0.5} \\
\midrule
Qwen3-VL-4B\textbf{*}
& \textbf{-}
& 36.2 & 13.1 & 16.6 & 7.0
& 36.1 & 8.9 & 10.2 & 3.8 \\
Qwen3-VL-8B\textbf{*}
& \textbf{-}
& 37.0 & 13.4 & 16.5 & 7.1
& 35.0 & 9.3 & 11.0 & 3.9 \\
Qwen3-VL-8B
& \textbf{196K}
& 42.8 & 22.4 & 29.6 & 15.3
& 41.4 & 18.2 & 19.6 & 12.8 \\
LLaVA-ST\textbf{*}
& \textbf{-}
& 45.1 & 14.3 & 18.3 & 7.4
& 43.0 & 11.4 & 13.9 & 5.8 \\
LLaVA-ST
& \textbf{196K}
& 48.4 & 22.5 & 29.0 & 17.4
& 45.3 & 15.4 & 19.2 & 10.6 \\
\rowcolor{gray!15}
DEViL\textbf{*}
& \textbf{101K}
& 44.4 & 25.2 & 32.8 & 22.7
& 43.2 & 18.2 & 22.8 & 15.1 \\
\rowcolor{gray!15}
DEViL
& \textbf{196K}
& \textbf{50.2} & \textbf{33.6} & \textbf{46.5} & \textbf{34.0}
& \textbf{48.5} & \textbf{28.8} & \textbf{38.7} & \textbf{28.2} \\
\bottomrule
\end{tabular}

\label{tab:vidstg_main}
\end{table*}

\section{Spatio-Temporal Training Data Details}
\label{sec:supp_training}

In the third training stage of DEViL,
we mix existing datasets with our self-collected data and automatically label them to train the model’s generalized spatio-temporal grounding capability.
Table \ref{tab:stvg_corpus_stats} summarizes the differences between these sources.
In the following,
we first describe the self-collected data and the auto-labeling pipeline,
and then introduce the instruction-style reformulation applied to the existing datasets.


\begin{figure}[t]
\centering

\includegraphics[width=1\linewidth]{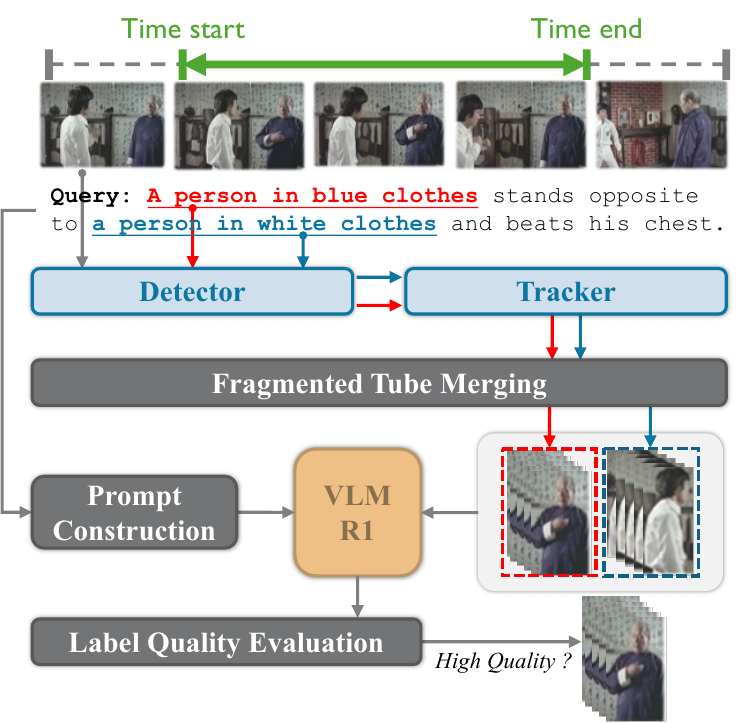}

\caption{\textbf{Auto-labeling process} used to translate temporal video grounding datasets to spatio-temporal video grounding datasets.}
\label{fig:autolabel}
\end{figure}

\noindent
\textbf{Self-collected Data for Training.}
Our self-collected dataset consists of two main parts: video segmentation datasets and temporal video grounding datasets.
First, following Sa2VA \cite{Sa2va}, we start from Ref-SAV \cite{Sa2va} derived from SA-V \cite{sam2}.
For each annotated object, we convert its segmentation masks into tight bounding boxes and align them with the annotated visible time span of the object.
This produces tube-level labels consisting of a start/end timestamp and a per-frame bounding box, providing dense spatial supervision for short clips.
Second,
we lift temporal video grounding (TVG) datasets, \textit{i.e.}, TACoS \cite{tacos}, ActivityNet Captions \cite{activitynet}, QVHighlights \cite{qvhighlight}, into spatio-temporal tubes.
Considering these datasets only provide temporal segments and text queries as the original annotations,
we build an automatic labeling pipeline to supplement the object tubes.



\begin{figure*}[t]
\centering
\includegraphics[width=1\linewidth]{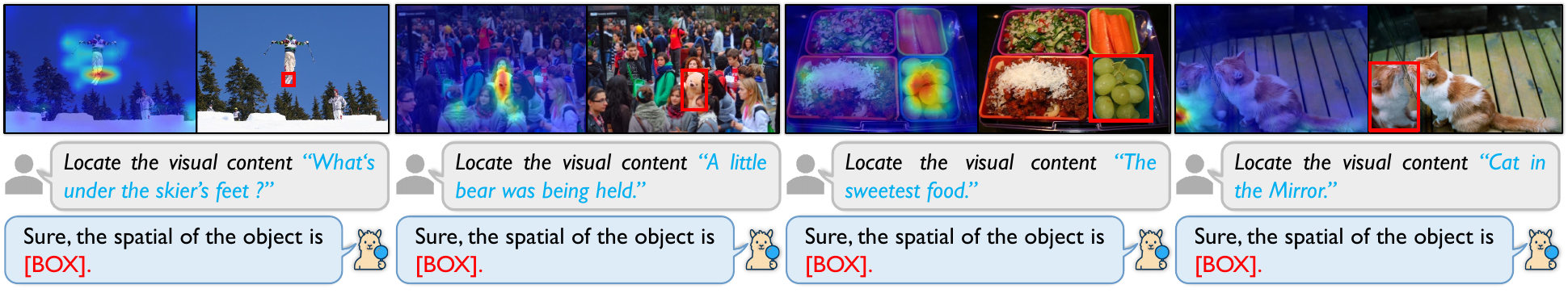}
\caption{Qualitative examples of image referring expression comprehension: given a natural-language query, our model predicts the spatial location of the target, and the overlaid heatmaps visualize attention between the \texttt{[BOX]}-induced RST/text feature and image features, which concentrates on the queried region.}

\label{fig:app_rec}
\end{figure*}

\begin{figure*}[t]
\centering
\includegraphics[width=1\linewidth]{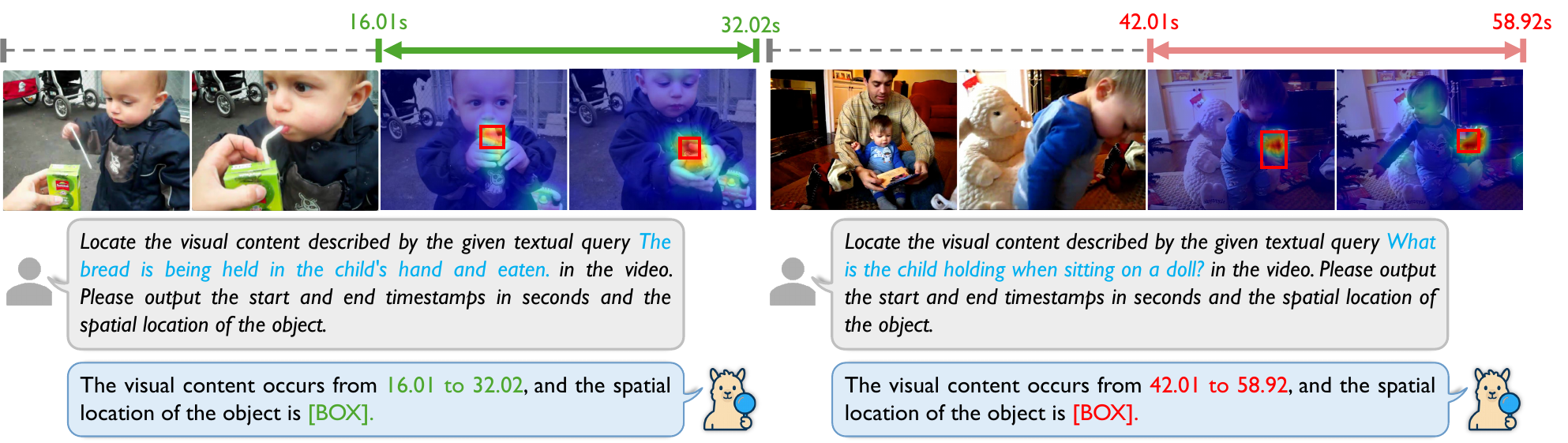}
\caption{Qualitative examples of spatio-temporal video grounding: given a natural-language query, our model predicts both the time span and the spatial location of the target, and the overlaid heatmaps visualize attention between the \texttt{[BOX]}-induced RST/text feature and video features, concentrating on the described event across time.}
\label{fig:app_stvg}
\end{figure*}

\noindent
\textbf{Auto-labeling Pipeline.}
As illustrated in Fig. \ref{fig:autolabel},
our auto-labeling pipeline for TVG datasets combines
(i) a strong open-vocabulary detector (MM-GroundingDINO \cite{mmgroundingdino}),
(ii) a referring-expression VLM (VLM-R1 \cite{vlmr1}), and
(iii) a visual tracker (SUTrack \cite{sutrack}).
Given a text query and its originally-annotated temporal interval, 
we first apply a detector–tracker pipeline to detect and track all objects mentioned in this query.
Next, to alleviate fragmented object tubes, we merge per-segment object tubes that belong to the same category and exhibit similar appearance but occur at different times,
forming complete object tubes as candidate results.
Then, leveraging VLM-R1 with powerful reasoning capability,
we select, from all candidates, the tube that best matches the text query.
Finally, since the selected tube may not fully cover the annotated interval,
we discard the produced annotation, i.e., the best tube,
if this tube overlaps with less than 50\% of the frames in the given interval.
Otherwise, we retain the selected tube as a pseudo annotation of spatio-temporal video grounding.



\noindent
\textbf{Instruction-Style Reformulation.}
We rewrite the supervision for all stages into concise question–answer instructions.

\medskip
\noindent\textbf{Stage~1 (Bridge).}\\[2pt]
\fcolorbox{gray!40}{gray!5}{%
  \parbox{0.97\linewidth}{\small\ttfamily
  Question: "<image> Locate the visual content described by the query <query></query> in the image."\\
  Answer:\ \ \ \ "The spatial location of the object is [BOX]."
  }%
}

\medskip
\noindent\textbf{Stage~2 (Alignment).}\\[2pt]
\fcolorbox{gray!40}{gray!5}{%
  \parbox{0.97\linewidth}{\small\ttfamily
  Question: "<video> Locate the visual content described by the query <query></query>. Output the start and end timestamps in seconds."\\
  Answer:\ \ \ \ "The visual content occurs from t1 to t2."
  }%
}

\medskip
\noindent\textbf{Stage~3 (Collaboration).}\\[2pt]
\fcolorbox{gray!40}{gray!5}{%
  \parbox{0.97\linewidth}{\small\ttfamily
  Question: "<video> Locate the visual content described by the query <query></query> in the video. Please output the start and end timestamps in seconds and the spatial location of the object."\\
  Answer:\ \ \ \ "The visual content occurs from t1 to t2, and the spatial location of the object is [BOX]."
  }%
}

\section{Additional Analysis}
\label{sec:more experiments}

We provide an additional analysis to further understand DEViL from the perspective of zero-shot generalization.

\noindent\textbf{Zero-shot STVG from the Self-collected Corpus.}
Tab.~\ref{tab:vidstg_main} reports VidSTG results under both zero-shot and supervised fine-tuning settings.
Generic MLLMs such as Qwen3-VL and LLaVA-ST show limited zero-shot grounding ability, but improve clearly after fine-tuning on the 196K STVG corpus.
In contrast, DEViL\textbf{*}, trained only on the 101K self-collected corpus, already achieves much stronger spatial grounding quality than the zero-shot baselines, especially on vIoU, vIoU@0.3, and vIoU@0.5.
After training on the full 196K corpus, DEViL further obtains the best performance on both declarative and interrogative splits, demonstrating the advantage of the detector-empowered design for spatio-temporal grounding.


\begin{figure*}[t]
\centering
\includegraphics[width=1\linewidth]{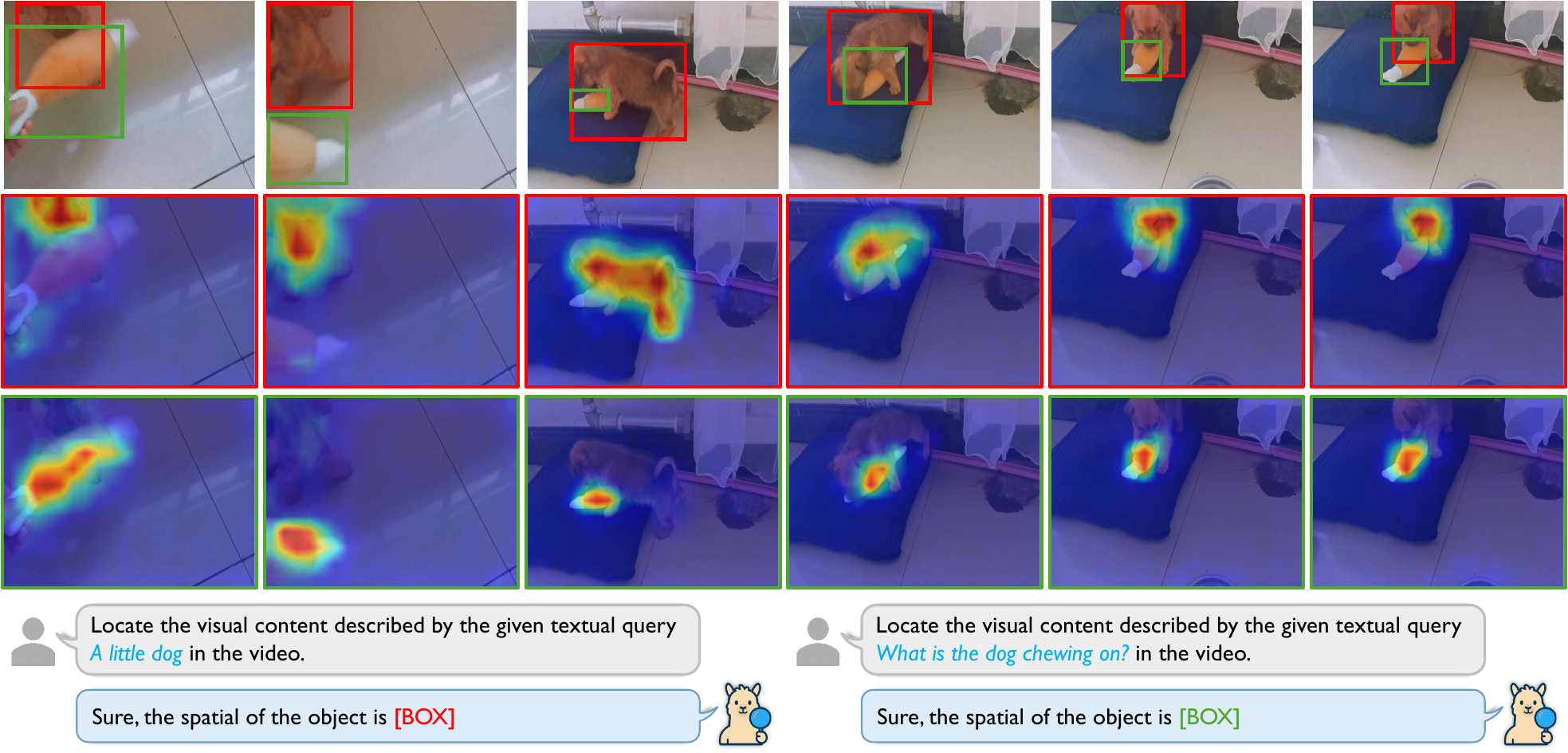}
\caption{Qualitative examples of spatial video grounding: given a natural-language query, our model predicts the frame-wise spatial location \texttt{[BOX]} of the target, and the overlaid heatmaps visualize attention between the \texttt{[BOX]}-induced RST/text feature and video features, focusing on the queried object across frames.}
\label{fig:app_svg}
\end{figure*}

\section{Failure Case Analysis}
\label{sec:failure_case}

Fig.~\ref{fig:failure_case} presents a challenging tracking failure case.
In this example, a small white dog is temporarily occluded by a larger dog.
During the occlusion, DEViL mistakenly shifts the predicted box to another object, indicating that the target identity is not fully preserved under severe visual interference.
Once the white dog reappears, however, DEViL is able to re-identify the target and recover correct grounding.
This example suggests that, although DEViL is reasonably robust to short-term occlusion, maintaining stable target association under heavy occlusion remains a challenging problem.

\begin{center}
\includegraphics[width=1\linewidth]{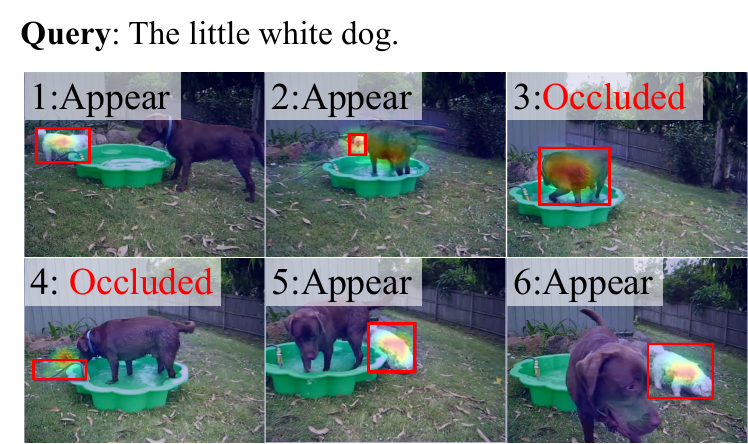}
\captionof{figure}{A hard failure case of spatio-temporal grounding. The small white dog is temporarily occluded by a larger dog. During the occlusion, DEViL mismatches the target box to another object, but re-identifies the correct target once it reappears.}
\label{fig:failure_case}
\end{center}

\section{Additional Qualitative Analysis}
\label{sec:visualized results}
\begin{figure*}[t]
\centering
\includegraphics[width=1\linewidth]{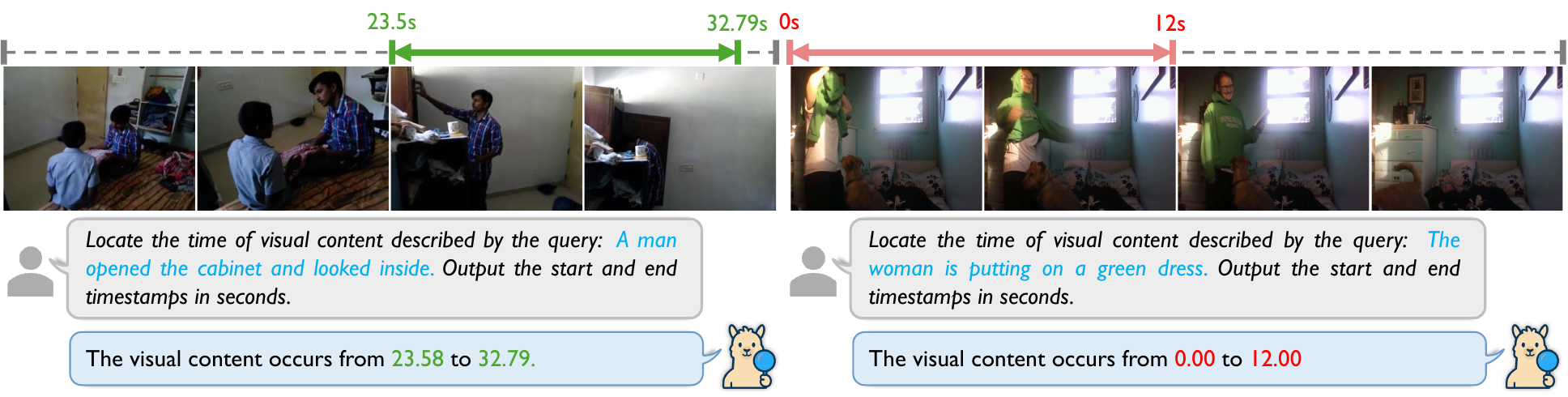}
\caption{Qualitative examples of temporal video grounding: given a language description, the model returns the start and end times of the corresponding event in the video.}
\label{fig:app_tvg}
\end{figure*}

\begin{figure*}[t]
\centering
\includegraphics[width=1\linewidth]{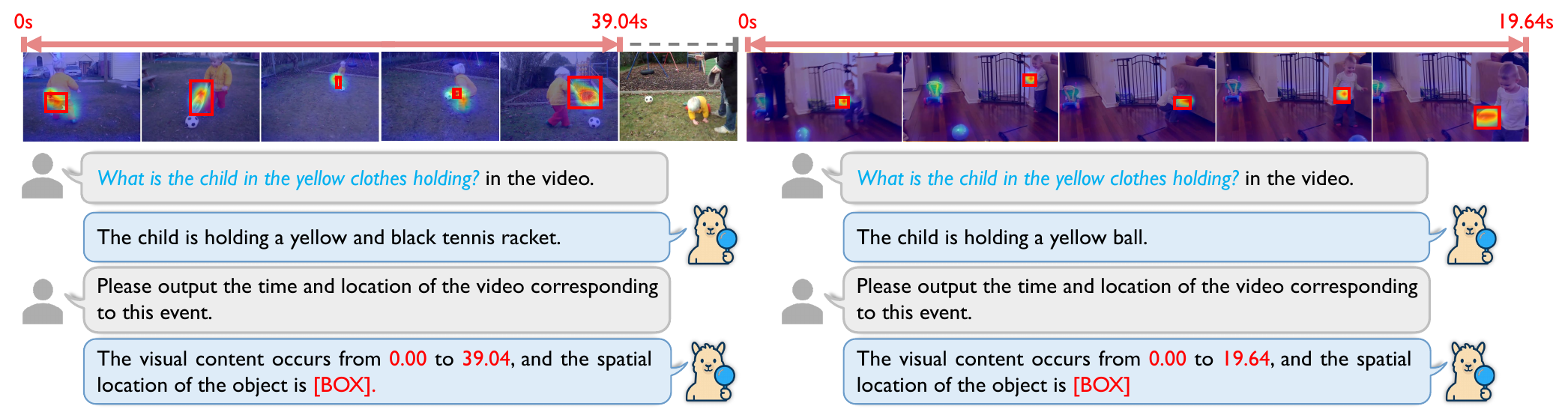}
\caption{Qualitative examples of grounded video question answering: given a natural-language question, our model first produces an answer and then predicts the time span and spatial location of the corresponding visual evidence, where the overlaid heatmaps visualize attention between the \texttt{[BOX]}-induced RST/text feature and video features, highlighting the evidence along the timeline.}
\label{fig:app_gvqa}
\end{figure*}

\begin{figure*}[t]
\centering
\includegraphics[width=1\linewidth]{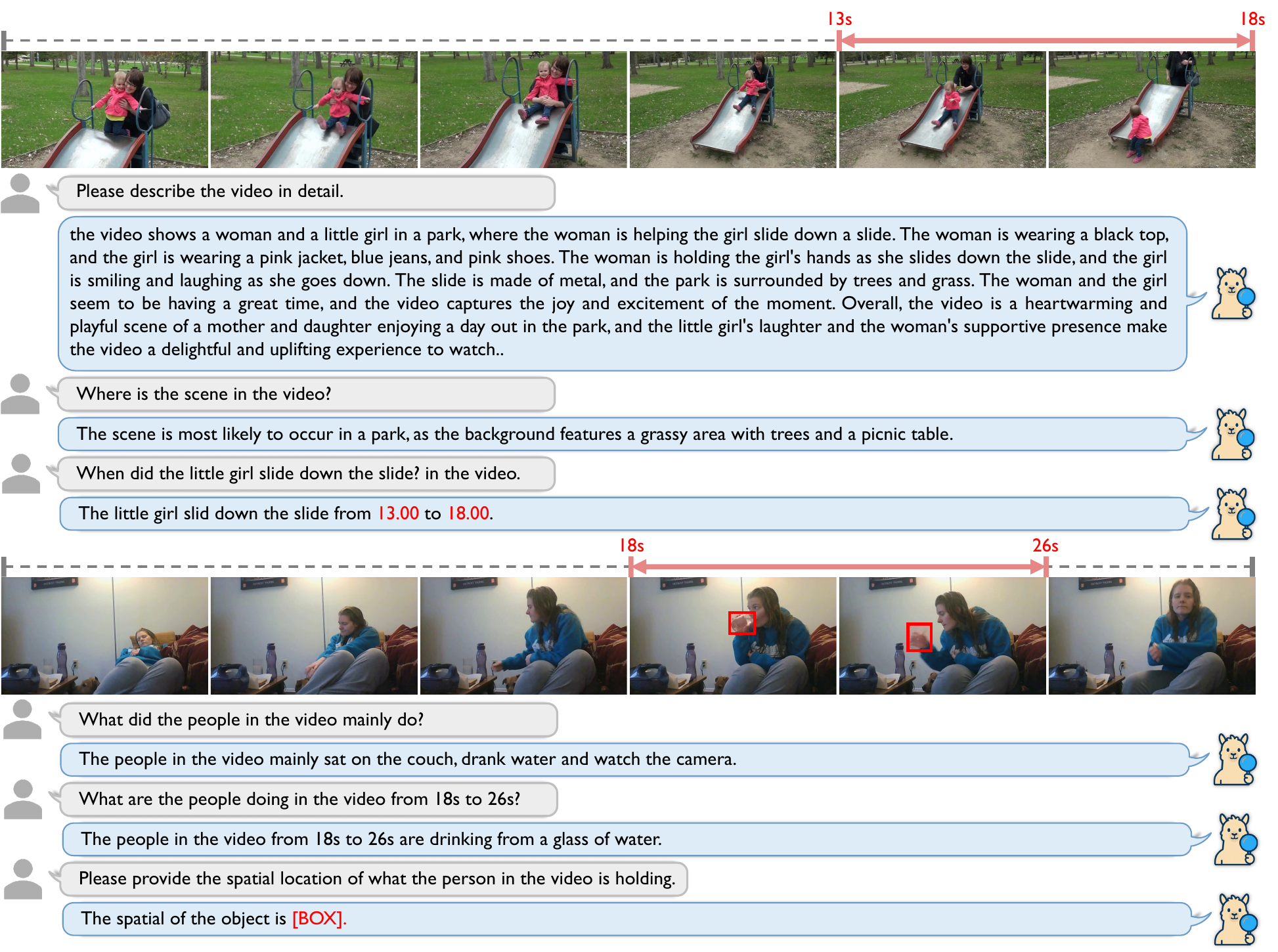}
\caption{Qualitative examples of multi-turn video conversation: our agent supports free-form descriptions, follow-up questions, and explicit temporal and spatial grounding within the same dialogue.}
\label{fig:app_conversation}
\end{figure*}

Figs.~\ref{fig:app_rec}–\ref{fig:app_conversation} present additional qualitative results across images and videos.
Fig.~\ref{fig:app_rec} shows image-level referring expression comprehension, where the predicted \texttt{[BOX]} and its RST-based attention align well with the queried region.
Figs.~\ref{fig:app_stvg} and \ref{fig:app_svg} illustrate spatio-temporal and spatial video grounding with attention consistently focusing on the target object.
Fig.~\ref{fig:app_tvg} highlights temporal event localization, while Fig.~\ref{fig:app_gvqa} shows grounded video QA supported by localized visual evidence.
Finally, Fig.~\ref{fig:app_conversation} demonstrates multi-turn video conversation, where our agent follows free-form instructions and provides explicit spatio-temporal grounding as interpretable evidence.

\section{Limitations and Future Work Discussion}
\label{sec:limitations}

Most existing spatio-temporal grounding (STVG) benchmarks adopt a \emph{single-target} setting, where each query corresponds to one dominant object tube. Consequently, DEViL is trained to emit a single RST and retrieve one tube per query, without explicitly modeling multiple entities or their roles. As future work, we plan to extend DEViL to \emph{multi-target} grounding by generating multiple entity-specific RSTs and adapting TTReg and the training corpus to maintain temporally consistent tubes in crowded scenes, bringing DEViL closer to video agents that reason over rich multi-entity interactions.

\ifdevilsupplementstandalone
\bibliographystyle{ACM-Reference-Format}
\bibliography{main}

\end{document}
\fi

\end{document}










\end{document}